\documentclass{article}
\usepackage{ijcai19}

\usepackage{times}
\usepackage{soul}
\usepackage{url}
\usepackage[hidelinks]{hyperref}
\usepackage[utf8]{inputenc}
\usepackage[small]{caption}
\usepackage{graphicx}
\usepackage{amsthm}
\usepackage{booktabs}
\usepackage{algorithm}
\usepackage{algorithmic}
\usepackage{amsmath,mathtools}
\usepackage{amssymb}
\usepackage{enumitem}
\usepackage{subfig}
\usepackage{multirow} 
\newcommand{\myeq}{\stackrel{\mathclap{\normalfont\mbox{def}}}{=}}
\newtheorem{lemma}{Lemma}
\renewcommand{\vec}[1]{\mathbf{#1}}
\DeclareMathOperator*{\argmax}{arg\,max}

\title{Multiple Partitions Aligned Clustering}

\author{
Zhao Kang$^1$\and
Zipeng Guo$^1$\and
Shudong Huang$^1$\and
Siying Wang$^1$\and\\
Wenyu Chen$^1$\and
Yuanzhang Su$^2$\And
Zenglin Xu$^1$\\
\affiliations
$^1$School of Computer Science and Engineering,\\
University of Electronic Science and Technology of China\\
$^2$School of Foreign Languages, University of Electronic Science and Technology of China\\
\emails
zkang@uestc.edu.cn, \{zpguo, huangsd, siyingwang\}@std.uestc.edu.cn, \{cwy, syz, zlx\}@uestc.edu.cn
}

\begin{document}

\maketitle

\begin{abstract}
Multi-view clustering is an important yet challenging task due to the difficulty of integrating the information from multiple representations. Most existing multi-view clustering methods explore the heterogeneous information in the space where the data points lie. Such common practice may cause significant information loss because of unavoidable noise or inconsistency among views. Since different views admit the same cluster structure, the natural space should be all partitions. Orthogonal to existing techniques, in this paper, we propose to leverage the multi-view information by fusing partitions. Specifically, we align each partition to form a consensus cluster indicator matrix through a distinct rotation matrix. Moreover, a weight is assigned for each view to account for the clustering capacity differences of views. Finally, the basic partitions, weights, and consensus clustering are jointly learned in a unified framework. We demonstrate the effectiveness of our approach on several real datasets, where significant improvement is found over other state-of-the-art multi-view clustering methods.
\end{abstract}

\section{Introduction}
As an important problem in machine learning and data mining, clustering has been extensively studied for many years \cite{jain2010data}. Technology advances have produced large volumes of data with multiple views. Multi-view features depict the same object from different perspectives, thereby providing complementary information. To leverage the multi-view information, multi-view clustering methods have drawn increasing interest in recent years \cite{chao2017survey}. Due to its unsupervised learning nature, multi-view clustering is still a challenging task. 
The key question is how to reach a consensus of clustering among all views. 

In the clustering field, two dominating methods are k-means \cite{jain2010data} and spectral clustering \cite{ng2002spectral}. Numerous variants of them have been developed over the past decades \cite{chen2013twkm,liu2018partition,yang2018fast,kang2018self}. Among them, some can tackle multi-view data, e.g., multi-view kernel k-means (MKKM) \cite{tzortzis2012kernel}, robust multi-view kernel k-means (RMKKM) \cite{cai2013multi}, Co-trained multi-view spectral clustering (Co-train) \cite{kumar2011cotrain}, Co-regularized multi-view spectral clustering (Co-reg) \cite{kumar2011co}. Along with the development of nonnegative matrix factorization (NMF) technique, multi-view NMF also gained a lot of attention. For example, a multi-manifold regularized NMF (MNMF) is designed to preserve the local geometrical structure of the manifolds for multi-view clustering \cite{zong2017multi}.  

Recently, subspace clustering method has shown impressive performance. Subspace clustering method first obtains a graph, which reveals the relationship between data points, then applies spectral clustering to achieve the embedding of original data, finally utilizes k-means to obtain the final clustering result \cite{elhamifar2013sparse,kang2019similarity}. Inspired by it, subspace clustering based multi-view clustering methods \cite{gao2015multi,zhang2017latent,huang2019auto} have become popular in recent years. For instance, Gao et al. proposed multi-view subspace clustering (MVSC) method \cite{gao2015multi}. In this approach, multiple graphs are constructed and they are forced to share the same cluster pattern. Therefore, the final clustering is a negotiated result and it might not be optimal. \cite{wang2016iterative} supposes that each graph should be close to each other. After obtaining graphs, their average is utilized to perform spectral clustering. The averaging strategy might be too simple to fully take advantage of heterogeneous information. Furthermore, it is a two-stage algorithm. The constructed graph might not be optimal for the subsequent clustering \cite{kang2017twin}.

By contrast, another class of graph-based multi-view clustering method learns a common graph based on adaptive neighbors idea \cite{nie2016parameter,zhan2017graph}. In specific, $x_i$ is connected to $x_j$ with probability $s_{ij}$. $s_{ij}$ should have a large value if the distance between $x_i$ and $x_j$ is small. Otherwise, $s_{ij}$ should be small. Therefore, obtained $s_{ij}$ is treated as the similarity between $x_i$ and $x_j$. In \cite{nie2016parameter}, each view shares the same similarity graph. Moreover, a weight for each view is automatically assigned based on loss value. Though this approach has shown its competitiveness, one shortcoming of it is that it fails to consider the flexible local manifold structures of different views. 

\begin{figure}[!htbp]
\includegraphics[width=.48\textwidth]{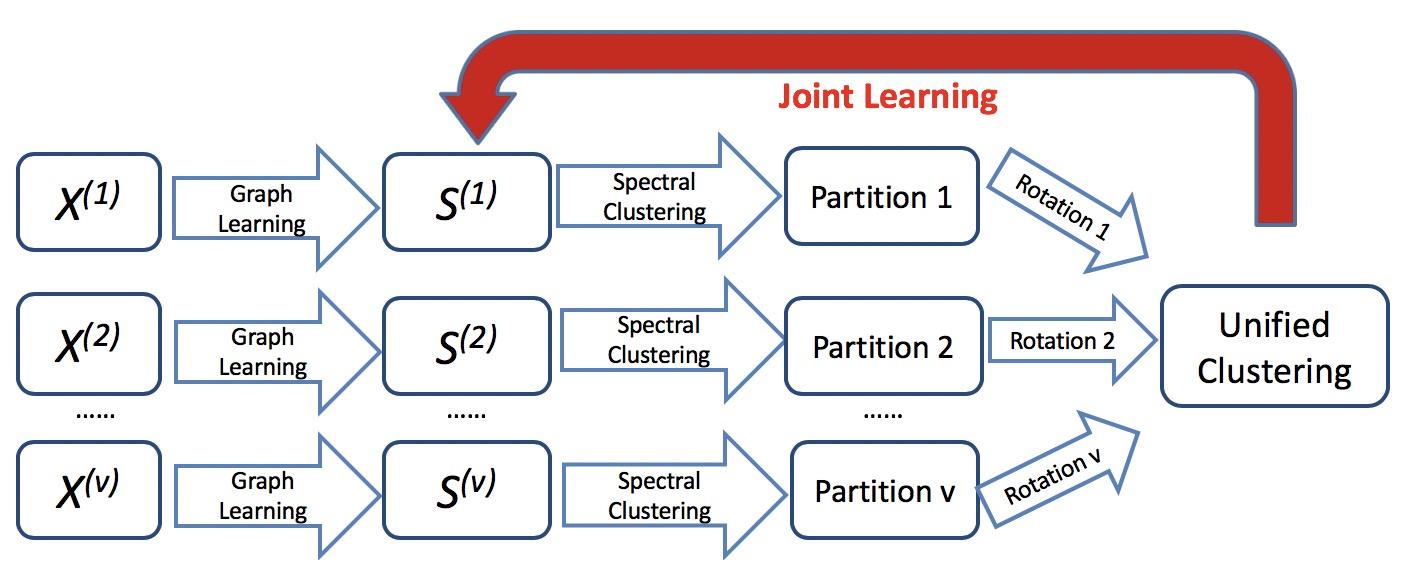}
\caption{Illustration of our mPAC. mPAC integrates graph learning, spectral clustering, and consensus clustering into a unified framework. \label{flow}}
\end{figure}

Although proved to be effective in many cases, existing graph-based multi-view clustering methods are limited in several aspects. First, they integrate the multi-view information in the feature space via some simple strategies. Due to the generally unavoidable noise in the data representation, the graphs might be severely damaged and cannot represent the true similarities among data points \cite{kang2019cyber}. It would make more sense if we directly reach consensus clustering in partition space where a common cluster structure is shared by all views, while the graphs might be quite different for different views. Hence, partitions from various views might be less affected by noise and easier to reach an agreement. Second, most existing algorithms follow a multi-stage strategy, which might degrade the final performance. For example, the learned graph might not be suitable for the subsequent clustering task. A joint learning method is desired for this kind of problem. 

Regarding the problems mentioned above, we propose a novel multiple Partitions Aligned Clustering (mPAC) method. Fig. \ref{flow} shows the idea of our approach. mPCA performs graph construction, spectral embedding, and partitions integration via joint learning. In particular, an iterative optimization strategy allows the consensus clustering to guide the graph construction, which later contributes to a new unified clustering. To sum up, we have our two-fold contributions as follows: 
\begin{itemize}[noitemsep]
\item{Orthogonal to existing multi-view clustering methods, we integrate multi-view information in partition space. This change in paradigm accompanies several benefits.}
\item{An end-to-end single stage model is developed to achieve from graph construction to final clustering. Especially, we assume that the unified clustering is reachable for each view through a distinct transformation. Moreover, the output of our algorithm is the discrete cluster indicator matrix, thus no more subsequent step is needed. }
\end{itemize}
\textbf{Notations} In this paper, matrices and vectors are represented by capital and lower-case letters, respectively. For $A=[a_{ij}]\in \mathbb{R}^{m\times n}$, $A_{i,:}$ and $A_{:,j}$ represents the $i$-th row and $j$-th column of $A$, respectively. The $\ell_2$-norm of vector $x$ is defined as $\|x\|=\sqrt{x^T \cdot x}$, where $T$ means transpose. $Tr(A)$ denotes the trace of $A$. $\|A\|_F=\sqrt{\sum_{i=1}^m\sum_{j=1}^n A_{ij}^2}$ denotes the Frobenius norm of $A$. Vector $\vec{1}$ indicates its elements are all ones. $I$ refers to the identity matrix with a proper size. $Ind \myeq \{Y\in\{0,1\}^{n\times c}|Y\vec{1}=\vec{1}\}$ represents the set of indicator matrices. We use the superscript $A^i$ or subscript $A_i$ to denote the $i$-th view of $A$ interchangeably when convenient.

\section{Subspace Clustering Revisited}
In general, for data $X\in\mathcal{R}^{m\times n}$ with $m$ features and $n$ samples, the popular subspace clustering method can be formulated as:
\vspace{-.2cm}
\begin{equation}
\min_S \|X-XS\|_F^2+\alpha \mathcal{R}(S)\quad s.t.\quad diag(S)=0,
\vspace{-.2cm}
\end{equation}  
where $\alpha>0$ is a balance parameter and $\mathcal{R}(S)$ is some regularization function, which varies in different algorithms \cite{peng2018connections}. For simplicity, we just apply the Frobenius norm in this paper. $diag(S)$ is the vector consists of diagonal elements of $S$. $S$ is treated as the affinity graph. Therefore, once $S$ is obtained, we can implement spectral clustering algorithm to obtain the clustering results, i.e.,
\vspace{-.2cm}
\begin{equation}
\min_{F} Tr(F^TLF)\hspace{0.2cm}s.t.\hspace{.15cm}F^TF=I,
\label{sc}
\vspace{-.2cm}
\end{equation}  

where $L\in\mathcal{R}^{n\times n}$ is the Laplacian of graph $S$ and $F\in\mathbb{R}^{n\times c}$ is the spectral embedding and $c$ is number of clusters. Graph Laplacian $L$ is defined by $L=D-S$, where $D$ is a diagonal matrix with $d_{ii}=\sum_j s_{ij}$. Since $F$ is not discrete, k-means is often used to recover the indicator matrix $Y\in Ind$.

When data of multiple views are available, Eq. (\ref{sc}) can be extended to this scenario accordingly. $X=[X^1;X^2;\cdots;X^v]\in \mathbb{R}^{m\times n}$ denotes the data with $v$ views, where $X^i\in\mathbb{R}^{m_i\times n}$ represents the $i$-th view data with $m_i$ features. Basically, most methods in the literature solve the following problem 
\vspace{-.2cm}
\begin{equation}
\min_{S, S^i} \sum\limits_i \|X^i-X^iS^i\|_F^2+\alpha G(S,S^i) \hspace{.13cm} s.t.\hspace{.13cm} diag(S^i)=0,
\label{msc}
\end{equation}
where $G$ represents some strategy to obtain a consensus graph $S$. For example, \cite{gao2015multi} enforces each graph to share the same $F$; \cite{wang2016iterative} penalizes the discrepancy between graphs, then their average is used as input to spectral clustering. 

We observe that there are several drawbacks shared by these approaches. First and foremost, they still lack an effective way to integrate multi-view knowledge while simultaneously considering the heterogeneity among views. Simply taking the average of graphs or assigning a unique spectral embedding is not enough to take full advantage of rich information. The graph representation itself might not be optimal to characterize the multi-view information. Secondly, they adopt a multi-stage approach. Since there is no mechanism to ensure the quality of learned graphs, this approach might lead to sub-optimal clustering results, which often occurs when noise exists. To address the above-mentioned challenging issues, we propose a multiple Partitions Aligned Clustering (mPAC) method. 
\section{Proposed Approach}
Unlike Eq.(\ref{msc}), which learns a unique graph based on multiple graphs $S^is$, we propose to learn a partition for each graph. In specific, we adopt a joint learning strategy and formulate our objective function as
\vspace{-.3cm}
\begin{multline}
\min_{S^i,F_i} \hspace{.2cm} \sum_{i=1}^{v} \Big\{\|X^i-X^iS^i\|_F^2+\alpha \|S^i\|_F^2+\beta Tr(F_i^TL^iF_i) \Big\}\\ s.t.\hspace{.2cm} diag(S^i)=0,\hspace{.06cm} F_i^T F_i=I.
\label{obj1}
\end{multline}
Next, we propose a way to fuse the multi-view information in the partition space. For multi-view clustering, a shared cluster structure is assumed. It is reasonable to assume a cluster indicator matrix $Y\in Ind$ for all views. Unfortunately, $F_i$'s elements are continuous. The discrepancy also exists among $F_i$'s. Thus, it is challenging to integrate multiple $F_i$s. To recover the underlying cluster $Y$, we assume that each partition is a perturbation of $Y$ and it can be aligned with $Y$ through a rotation \cite{kang2018unified,nie2018multiview}. Mathematically, it can be formulated as
\vspace{-.5cm}
\begin{equation}
\min_{Y,R_i}  \sum_{i=1}^{v}\|Y-F_iR_i\|_F^2\hspace{.2cm} s.t.\hspace{.2cm} Y\in Ind,\hspace{.1cm} R_i^TR_i=I,
\label{obj2}
\end{equation}
where $R_i$ represents an orthogonal matrix. Eq. (\ref{obj2}) treats each view equally. As shown by many researchers, it is necessary to distinguish their contributions. Therefore, we introduce a weight parameter $w_i$ for view $i$.
Deploying a unified framework, we eventually reach our objective for mPAC as
\vspace{-.2cm}
\begin{equation}
\begin{split}
\min_{S^i,F_i,Y,w_i,R_i} \hspace{.2cm} &\sum_{i=1}^{v} \Big\{\|X^i-X^iS^i\|_F^2+\alpha \|S^i\|_F^2+\\
&\beta Tr(F_i^TL^iF_i)+\frac{\gamma}{w_i}\|Y-F_iR_i\|_F^2 \Big\}\\
&\hspace{-.86cm}s.t.\hspace{.18cm} diag(S^i)=0,\hspace{.06cm} F_i^T F_i=I,\hspace{.06cm}Y\in Ind,\\
&\hspace{.06cm} R_i^T R_i=I,\hspace{.06cm} w_i\ge 0, \hspace{.06cm} w\vec{1}=1.
\label{obj}
\end{split}
\end{equation}
We can observe that the proposed approach is distinct from other methods in several aspects:

\begin{itemize}[noitemsep]
\item{Orthogonal to existing multi-view clustering techniques, Eq. (\ref{obj}) integrates heterogeneous information in partition space. Considering that a common cluster structure is shared by all views, it would be natural to perform information fusion based on partitions.   }
\item{Generally, learning with multi-stage strategy often leads to sub-optimal performance. We adopt a joint learning framework. The learning of similarity graphs, spectral embeddings, view weights, and unified cluster indicator matrix is seamlessly integrated together. }
\item{$Y$ is the final discrete cluster indicator matrix. Hence, discretization procedure is no longer needed. This eliminates the k-means post-processing step, which is sensitive to initialization. With input $X$, (\ref{obj}) will output the final discrete $Y$. Thus, it is an end-to-end single-stage learning problem.}
\item{Multiple graphs are learned in our approach. Hence, the local manifold structures of each view are well taken care of.}
\item{As a matter of fact, Eq. (\ref{obj}) is not a simple unification of the pipeline of steps and it attempts to learn graphs with optimal structure for clustering. According to the graph spectral theory, the ideal graph is $c$-connected if there are $c$ clusters \cite{kang2018unified}. In other words, the Laplacian matrix $L$ has $c$ zero eigenvalues $\sigma_i$s. Approximately, we can minimize $\sum_{i=1}^c \sigma_i$, which is equivalent to $\min\limits_{F^TF=I} Tr(F^TLF)$. Hence, the third term in Eq. (\ref{obj}) ensures that each graph $S^i$ is optimal for clustering.}
\end{itemize}

%

\section{Optimization Methods}
To handle the objective function in Eq. (\ref{obj}), we apply an alternating minimization scheme to solve it. 
\subsection{Update $S^i$ for Each View}
By fixing other variables, we solve $S^i$ according to
\vspace{-.2cm}
\begin{multline}
\min_{S^i} \hspace{.1cm} \sum_{i=1}^{v} \Big\{\|X^i-X^iS^i\|_F^2+\alpha \|S^i\|_F^2+
\beta Tr(F_i^TL^iF_i)\Big\}\hspace{-.16cm}\\
\hspace{.88cm}s.t.\hspace{.18cm} diag(S^i)=0.
\end{multline}
It can be seen that each $S^i$ is independent from other views. Therefore, we can solve each view separately. 
To simplify the notations, we ignore the view index tentatively. Note that $L$ is a function of $S$ and $Tr(F^TLF)=\sum_{ij} \frac{1}{2}\|F_{i,:}-F_{j,:}\|^2S_{ij}$. Equivalently, we solve
\begin{equation}
\min_{S_{:,i}} \hspace{.1cm} \|X_{:,i}-XS_{:,i}\|^2+\alpha S_{:,i}^TS_{:,i}+\frac{\beta}{2}h_i^TS_{:,i},
\end{equation}
where $h_i\in\mathcal{R}^{n\times 1}$ with the $j$-th component defined by $h_{ij}=\|F_{i,:}-F_{j,:}\|^2$. By setting its first-order derivative to zero, we obtain
\vspace{-.2cm}
\begin{equation}
S_{:,i}=(\alpha I+X^TX)^{-1}\Big[(X^TX)_{i,:}-\frac{\beta h_i}{4}\Big].
\label{updates}
\end{equation}
\subsection{Update $F_i$ for Each View}
Similarly, we drop all unrelated terms with respect to $F_i$ and ignore the view indexes. It yields,
\begin{equation}
\min_F   \beta Tr(F^TLF)+\frac{\gamma}{w_i}\|Y-FR\|_F^2 \hspace{.2cm}s.t.\hspace{.1cm} F^TF=I.
\label{updatef}
\end{equation}
This sub-problem can be efficiently solved based on the method developed in \cite{wen2013feasible}.
\subsection{Update $R_i$ for Each View}
With respect to $R_i$, the objective function is additive. We can solve each $R_i$ individually. Specifically,
\begin{equation}
\min_R  \|Y-FR\|_F^2 \hspace{.2cm}s.t.\hspace{.1cm} R^TR=I.
\label{updateR}
\end{equation}
\begin{lemma}
For problem 
\begin{equation}
\min\limits_{R^TR=I} \|Y-FR\|_F^2, 
\end{equation}
its closed-form solution is $R^*=UV^T$, where $U$, $V$ are the left and right unitary matrix of the SVD decomposition of $F^TY$, respectively \cite{schonemann1966generalized}.
\end{lemma}
\subsection{Update $Y$}
For $Y$, we get
\begin{equation}
\min_Y  \sum\limits_{i=1}^{v}\frac{1}{w_i}\|Y-F_iR_i\|_F^2 \hspace{.2cm} s.t.\hspace{.1cm} Y\in Ind.
\end{equation}
Let's unfold above objective function, we have
\begin{eqnarray*}
\begin{aligned}
&\sum\limits_{i=1}^{v}\frac{1}{w_i}\|Y-F_iR_i\|_F^2\\
=&\sum\limits_{i=1}^{v}\frac{1}{w_i}(\|Y\|_F^2+\|F_iR_i\|_F^2)-\sum_{i=1}^{v}\frac{2}{w_i}Tr(Y^TF_iR_i)\\
=&\sum\limits_{i=1}^{v}\frac{n+c}{w_i}-2Tr\Big(Y^T(\sum\limits_{i=1}^{v}\frac{F_iR_i}{w_i})\Big).
\end{aligned}
\end{eqnarray*}
Thus, we can equivalently solve
\begin{equation}
\max_{Y\in Ind} \hspace{.1cm}Tr\Big(Y^T(\sum\limits_{i=1}^{v}\frac{F_iR_i}{w_i})\Big).
\end{equation}
It admits a closed-form solution, that is, $\forall i=1,\cdots,n$, 
\begin{equation}
Y_{ij}=
\begin{cases}
1&j=\argmax\limits_k [\sum\limits_{i=1}^{v}\frac{F_i R_i}{w_i}]_k,\\
0&\text{otherwise.}
\end{cases}
\label{updateY}
\end{equation}

\subsection{Update $w_i$ for Each View}
Let's denote $\|Y-F_iR_i\|_F$ as $q_i$, then this subproblem can be expressed as 
\vspace{-.2cm}
\begin{equation}
\min_{w_i\geq 0, w\vec{1}=1} \sum\limits_{i=1}^{v}\frac{q_i^2}{w_i}.
\end{equation}
Based on Cauchy-Schwarz inequality, we have
\begin{equation}
\sum\limits_{i=1}^{v}\frac{q_i^2}{w_i}=\Big(\sum\limits_{i=1}^{v}\frac{q_i^2}{w_i}\Big) \Big(\sum\limits_{i=1}^vw_i\Big)\geq \Big(\sum\limits_{i=1}^v q_i\Big)^2.
\end{equation}
The minimum, which is a constant, is achieved when $\sqrt{w_i}\propto\frac{q_i}{\sqrt{w_i}}$. Thus, the 
optimal $w$ is given by, $\forall i=1,\cdots,v$, 
\vspace{-.2cm}
\begin{equation}
w_i=\frac{q_i}{\sum\limits_{i=1}^{v} q_i}.
\label{weight}
\vspace{-.2cm}
\end{equation}.

For clarity, we summarize the algorithm\footnote{Our code is available: https://github.com/sckangz/mPAC} to solve Eq. (\ref{obj}) in Algorithm 1. 
\begin{algorithm}[!htbp]
\caption{Optimization for mPAC}
\label{alg1}
\scriptsize
 {\bfseries Input:} Multiview matrix $X^1, \cdots, X^v$, cluster number $c$, parameters $\alpha$, $\beta$, $\gamma$.\\
{\bfseries Output:} $Y$.\\
{\bfseries Initialize:} Random $Y$ and $F_i$, $R_i=I$, $w_i=1/v, \forall i=1,\cdots,v$.\\
 {\bfseries REPEAT}
\begin{algorithmic}[1]
\STATE\hspace{.4cm} \textbf{for} view 1 to $v$ \textbf{do}
\STATE \hspace{.6cm} Update each column of $S$ according to (\ref{updates});\\
\STATE \hspace{.6cm} Solve the subproblem (\ref{updatef});\\
\STATE \hspace{.6cm} Solve the subproblem (\ref{updateR});\\
\STATE \hspace{.4cm} \textbf{end for}
\STATE \hspace{.4cm}  Update $Y$ according to (\ref{updateY});
\STATE  \hspace{.4cm} Update $w_i$ via (\ref{weight}) for each view.
\end{algorithmic}
\textbf{ UNTIL} {stopping criterion is met}
\end{algorithm}

\section{Experiments}
\begin{table}[!tbp]
\begin{center}

\label{datasets} \scalebox{.9}{
\begin{tabular}{cllll}
\hline
{Data} & {Handwritten} &{Caltech7} &  {Caltech20} & {BBCSport}\\\hline
View \# &6&6&6&4\\
Points& 2000 & 1474  & 2386 & 116\\
Cluster \#&10 & 7  & 20 & 5 \\
\hline
\end{tabular}}
\end{center}
\caption{Description of the data sets. \label{data}}
\end{table}
\subsection{Experimental Setup}
We conduct experiments on four benchmark data sets: BBCSport, Caltech7, Caltech20, Handwritten 
Numerals. Their statistics information is summarized in Table \ref{data}.
We compare the proposed mPAC with several state-of-the-art methods from different categories, including Co-train~\cite{kumar2011cotrain}, Co-reg~\cite{kumar2011co}, MKKM~\cite{tzortzis2012kernel}, RMKM~\cite{cai2013multi}, MVSC~\cite{gao2015multi}, MNMF~\cite{zong2017multi}, parameter-free auto-weighted multiple graph learning (AMGL) \cite{nie2016parameter}.
Furthermore, the classical k-means (KM) method with concatenated features (i.e., all features, AllFea in short) is included as a baseline. That is to say, all views are of the same importance. Following \cite{huang2018self}, all values of each view are normalized into range $[-1,1]$. To achieve a comprehensive evaluation, we apply five widely-used metrics to examine the effectiveness of our method: F-score, precision, Recall, Normalized Mutual Information (NMI), and Adjusted Rand Index (ARI). We initialize our algorithm by using the results from \cite{nie2016constrained}.
 
\begin{table*}[!htbp]	
			\renewcommand{\arraystretch}{1.1}
	\centering
			\setlength{\tabcolsep}{7pt}{

				\begin{tabular}{|c |c | c| c| c|c|}
					\hline
					Method& F-score& Precision& Recall& NMI& ARI\\
					\hline	
KM(AllFea)&0.3834(0.0520)&0.2345(0.0463)&0.6616(0.2161)&0.1701(0.0763)&0.1561(0.0863)\\
					Co-train&0.3094(0.0107)&0.2348(0.0034)&0.4556(0.0398)&0.1591(0.0160)&0.1144(0.0064)\\
					Co-reg&0.3116(0.0305)&0.2337(0.0053)&0.4879(0.1173)&0.1599(0.0192)&0.1166(0.0090)\\
					MKKM&0.3779(0.0162)&0.2359(0.0156)&0.7679(0.1402)&0.1160(0.0392)&0.1248(0.0309)\\
					RMKM&0.3774(0.0167)&0.2476(0.0113)&0.8416(0.1563)&0.1754(0.0259)&0.1100(0.0200)\\
					MVSC&0.3540(0.0270)&0.2459(0.0406)&0.7017(0.0801)&0.1552(0.0812)&0.1292(0.0666)\\
					
					MNMF&0.3755(0.0307)&0.2685(0.0117)&\textbf{0.8558(0.1261)}&0.2576(0.0614)&0.1274(0.0515)\\
					AMGL&0.3963(0.0167)&0.2801(0.0226)&0.6976(0.0971)&0.2686(0.0419)&0.0785(0.0399)\\
mPAC& \textbf{0.6780}&	\textbf{0.7500}&	0.6187&	\textbf{0.6146}&	\textbf{0.5617}\\
					\hline			
			\end{tabular}
	
	}
		\caption{Clustering performance on BBCSport data.\label{bbc}}
			\vspace{.55cm}
			\renewcommand{\arraystretch}{1.1}
			
			\setlength{\tabcolsep}{7pt}{
		
				\begin{tabular}{|c |c | c| c| c|c|}
					\hline
					Method& F-score& Precision& Recall& NMI& ARI\\
KM(AllFea)&0.4688(0.0327)&0.7868(0.0080)&0.3618(0.0371)&0.4278(0.0120)&0.3172(0.0297)\\
					Co-train&0.4678(0.0172)&0.7192(0.0136)&0.3550(0.0168)&0.3235(0.0226)&0.3342(0.0157)\\
					Co-reg&0.4981(0.0092)&0.7014(0.0076)&0.3622(0.0098)&0.3738(0.0061)&0.2894(0.0046)\\
					MKKM&0.4804(0.0059)&0.7659(0.0178)&0.3663(0.0040)&0.4530(0.0132)&0.3053(0.0096)\\
					RMKM&0.4514(0.0409)&0.7491(0.0277)&0.3236(0.0376)&0.4220(0.0197)&0.2865(0.0429)\\
					MVSC&0.3341(0.0102)&0.5387(0.0271)&0.2427(0.0130)&0.1938(0.0185)&0.1242(0.0140)\\
					
					MNMF&0.4414(0.0303)&\textbf{0.7587(0.0330)}&0.3115(0.0262)&0.4111(0.0175)&0.3456(0.0576)\\
	AMGL&0.6422(0.0139)&0.6638(0.0125)&0.6219(0.0164)&0.5711(0.0149)&0.4295(0.0208)\\				
mPAC&	\textbf{0.6763}&	0.6306&	\textbf{0.7292}&	\textbf{0.5741}&	\textbf{0.4963}\\	
\hline
						
			\end{tabular}
	}
	\caption{Clustering performance on Caltech7 data.\label{caltech7}}
			\vspace{.55cm}	
			\renewcommand{\arraystretch}{1.1}
			
			\setlength{\tabcolsep}{7pt}{
		
				\begin{tabular}{|c |c | c| c| c|c|}
					\hline
					Method& F-score& Precision& Recall& NMI& ARI\\
					\hline					
					KM(AllFea)&0.3697(0.0071)&0.6235(0.0212)&0.2583(0.0095)&0.5578(0.0133)&0.2850(0.0063)\\
					Co-train&0.3750(0.0287)&0.6375(0.0253)&0.2749(0.0238)&0.4895(0.0117)&0.3085(0.0281)\\
					Co-reg&0.3719(0.0087)&0.6245(0.0137)&0.2882(0.0070)&0.5615(0.0042)&0.2751(0.0084)\\
	MKKM&0.3583(0.0114)&\textbf{0.6724(0.0158)}&0.2865(0.0092)&0.5680(0.0142)&0.3039(0.0110)\\
					RMKM&0.3955(0.0113)&0.6307(0.0144)&0.2712(0.0096)&0.5899(0.0092)&0.2952(0.0112))\\
					MVSC&0.5417(0.0239)&0.4100(0.0245)&0.7994(0.0110)&0.4875(0.0113)&0.3800(0.0246)\\
					MNMF&0.3643(0.0157)&0.6509(0.0119)&0.2530(0.0136)&0.5367(0.0132)&0.3128(0.0042)\\
	AMGL&0.4017(0.0248)&0.3503(0.0479)&0.4827(0.0450)&0.5656(0.0387)&0.2618(0.0453)\\				
					mPAC& 	\textbf{0.5645}&	0.4350&	\textbf{0.8035}&	\textbf{0.5986}&	\textbf{0.5083}\\	
\hline
						
			\end{tabular}
		}
	\caption{Clustering performance on Caltech20 data.\label{caltech20}}
%
		\vspace{.55cm}		
			\renewcommand{\arraystretch}{1.1}
			
			\setlength{\tabcolsep}{7pt}{
			
				\begin{tabular}{|c |c | c| c| c|c|}
					\hline
					Method& F-score& Precision& Recall& NMI& ARI\\
					\hline					
%
%
%
%
%
%
					KM(AllFea)&0.6671(0.0105)&0.6550(0.0154)&0.6889(0.0180)&0.7183(0.0106)&0.6443(0.0122)\\
					
					
					Co-train&0.6859(0.0172)&0.6634(0.0281)&0.7109(0.0252)&0.7222(0.0149)&0.6498(0.0227)\\
					
					Co-reg&0.6840(0.0269)&0.6360(0.0336)&0.6413(0.0198)&0.7583(0.0197)&0.6266(0.0314)\\
					
					MKKM&0.6756(0.0000)&0.6501(0.0000)&0.7050(0.0000)&0.7526(0.0000)&0.7009(0.0000)\\
					
					RMKM&0.6542(0.0258)&0.6218(0.0350)&0.6915(0.0158)&0.7431(0.0209)&0.6013(0.0300)\\
					
					MVSC&0.6753(0.0335)&0.6193(0.0537)&0.7537(0.0215)&0.7566(0.0186)&0.6079(0.0419)\\
					
					MNMF& 0.7068(0.0272)&0.6957(0.0294)&0.7183(0.0250)&0.7431(0.0227)&0.6407(0.0056)\\
					AMGL&0.7404(0.1070)&0.6650(0.1372)&\textbf{0.8457(0.0560)}&\textbf{0.8392(0.0543)}&0.7066(0.1235)\\
			mPAC&	\textbf{0.7473}&	\textbf{0.7348}&	0.7200&	0.7370&	\textbf{0.7069}\\

\hline
						
			\end{tabular}
		}
\caption{Clustering performance on Handwritten numerals data.\label{handwritten}}

	\end{table*}

\begin{figure}[!htbp]
\centering
\includegraphics[angle=270,width=.465\textwidth]{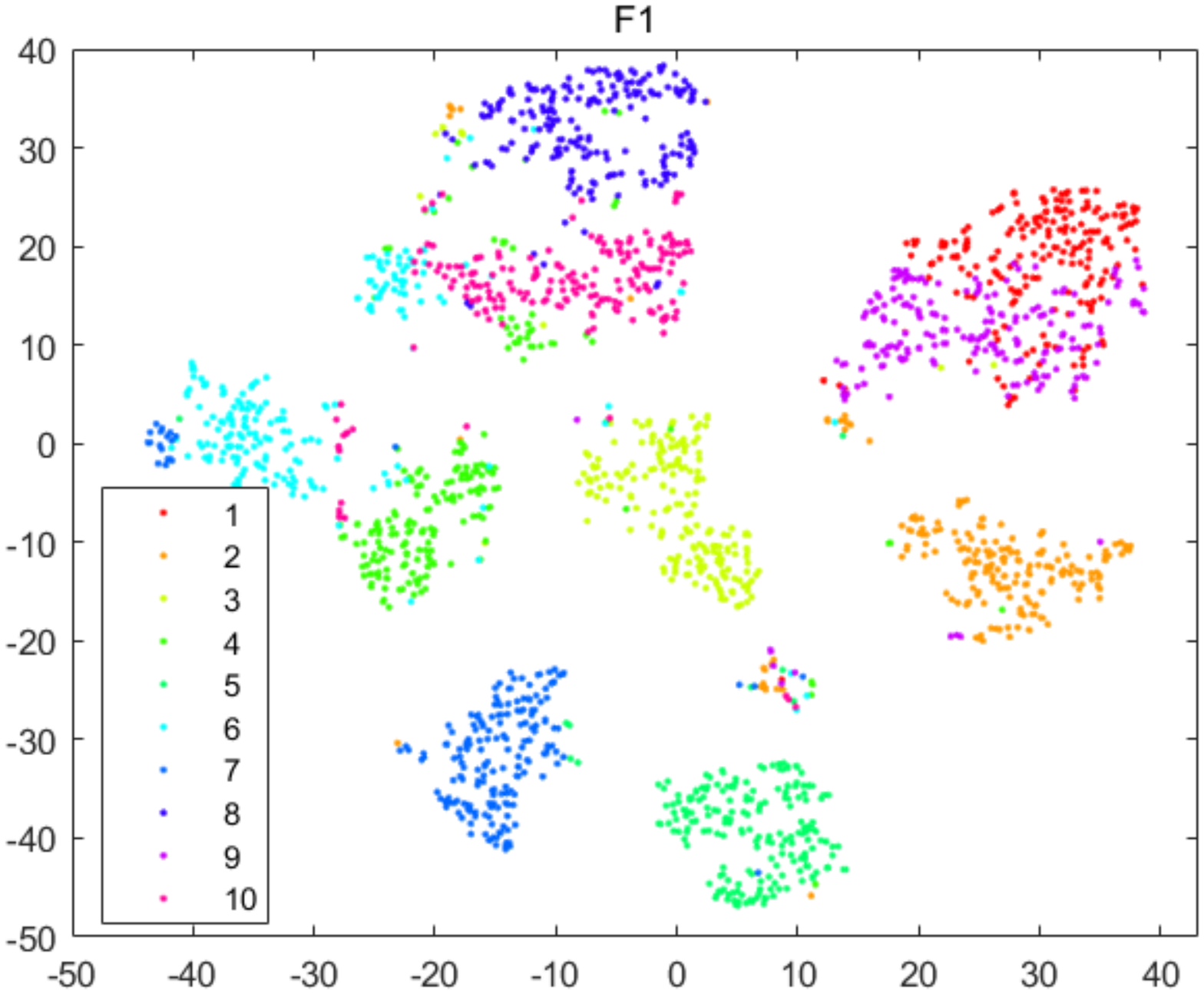}\vspace{-.6cm}\\
\includegraphics[angle=270,width=.465\textwidth]{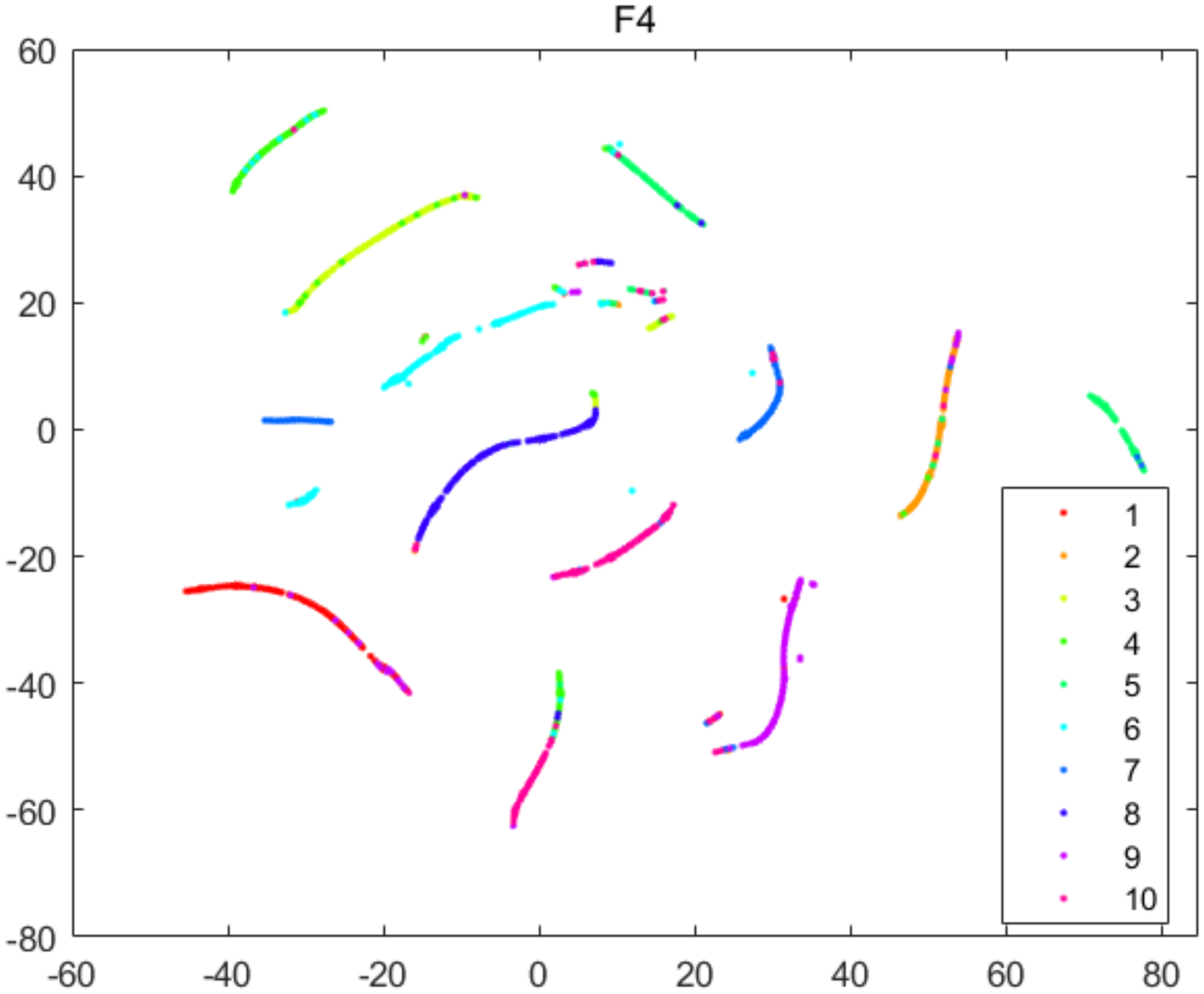}\vspace{-.6cm}\\
\includegraphics[angle=270,width=.465\textwidth]{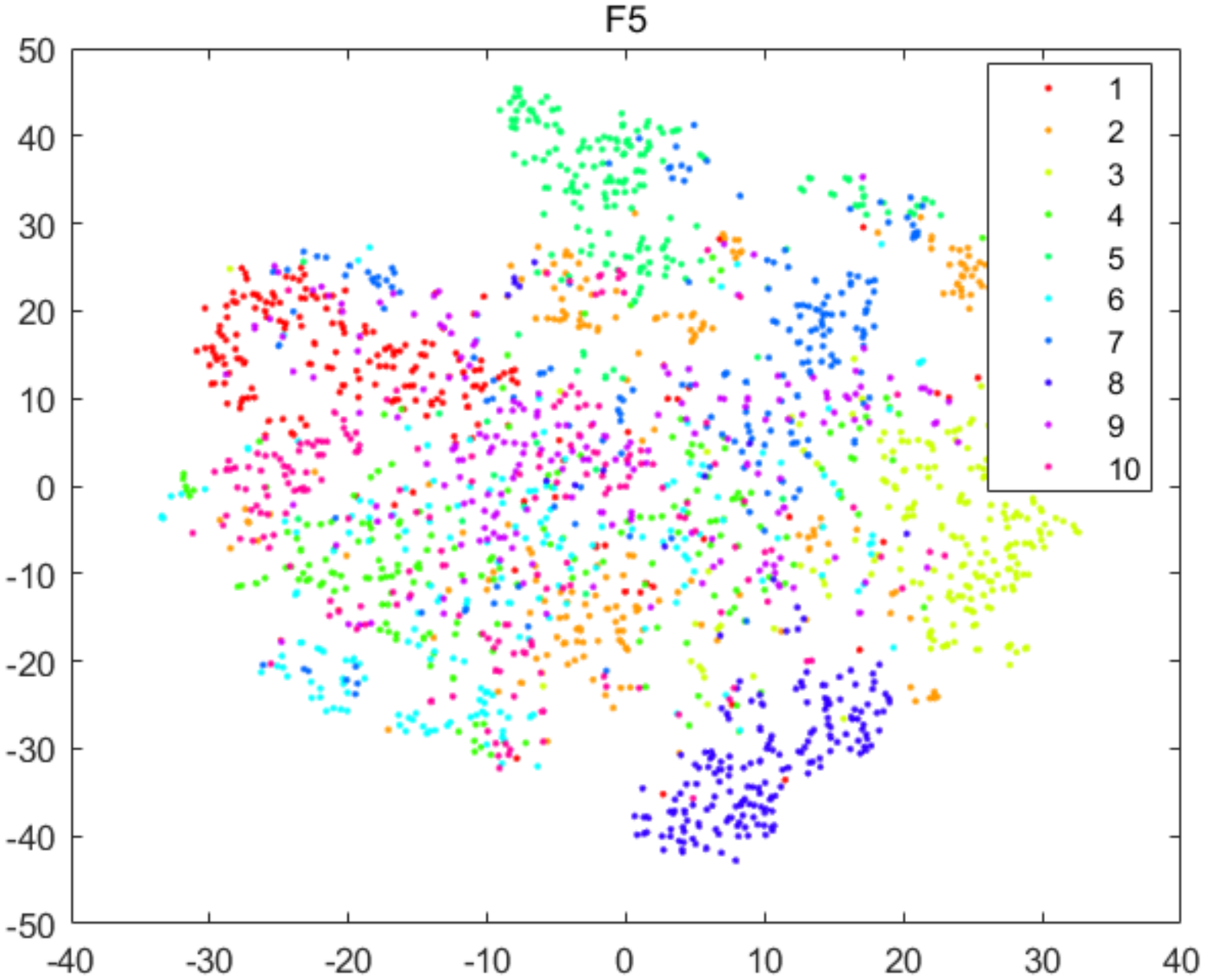}\vspace{-.6cm}\\
\includegraphics[angle=270,width=.465\textwidth]{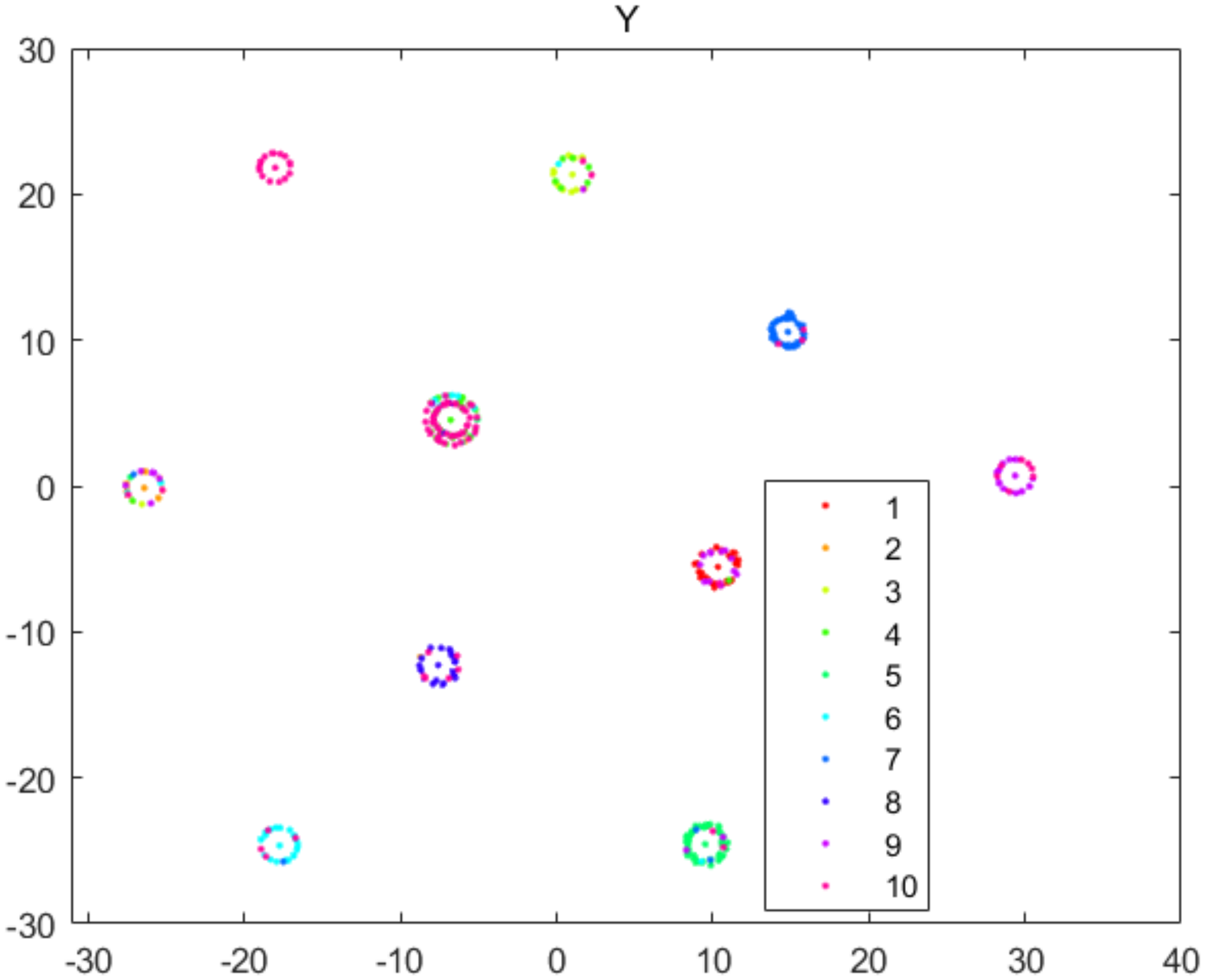}
\caption{Some clustering results of the Handwritten Numerals data set.\label{partition}}
\end{figure}

\begin{figure}[h]
\centering
\subfloat[$\gamma=10^{-6}$]{\includegraphics[width=.4\textwidth]{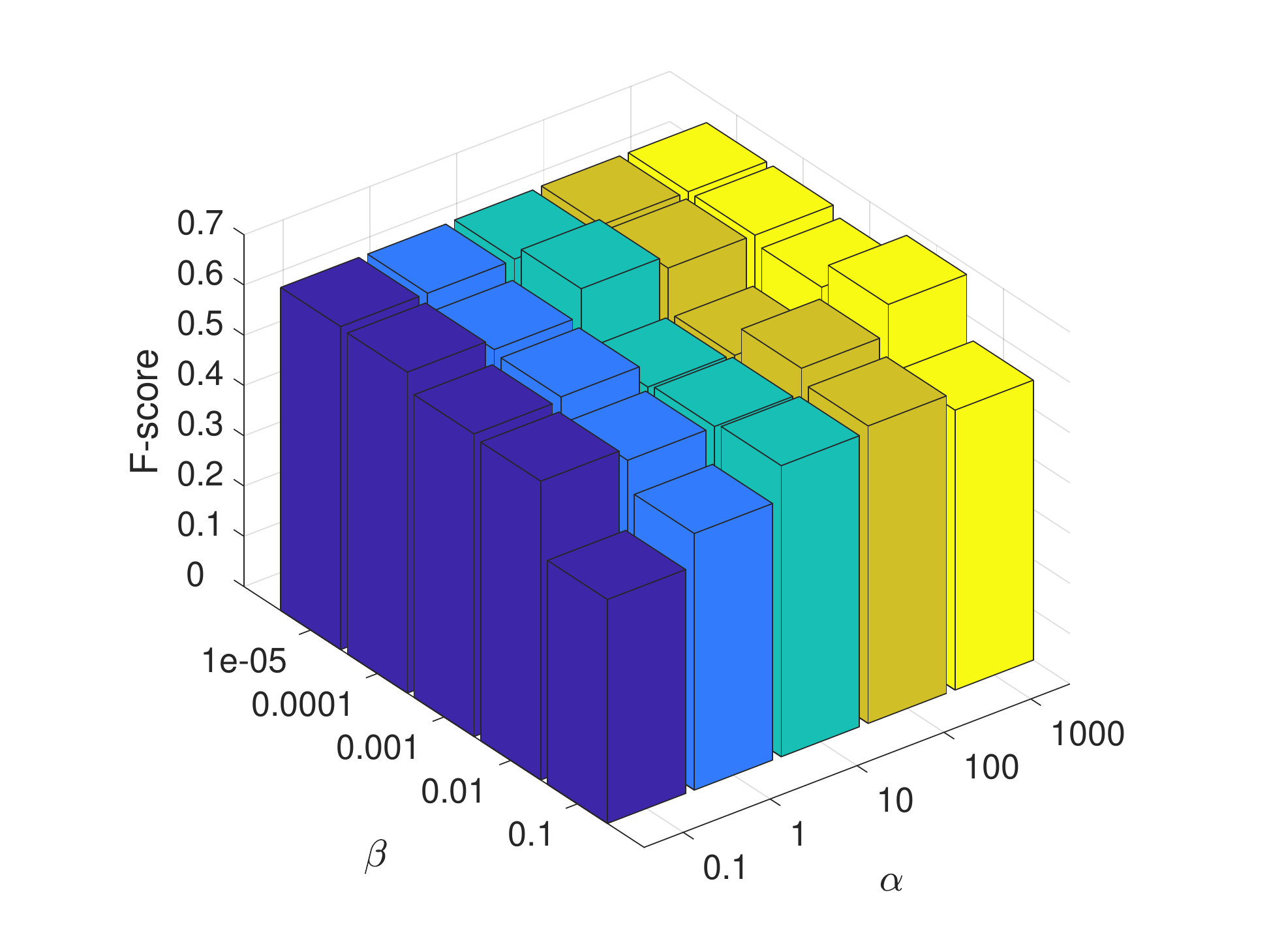}}\\
\subfloat[$\gamma=10^{-3}$]{\includegraphics[width=.4\textwidth]{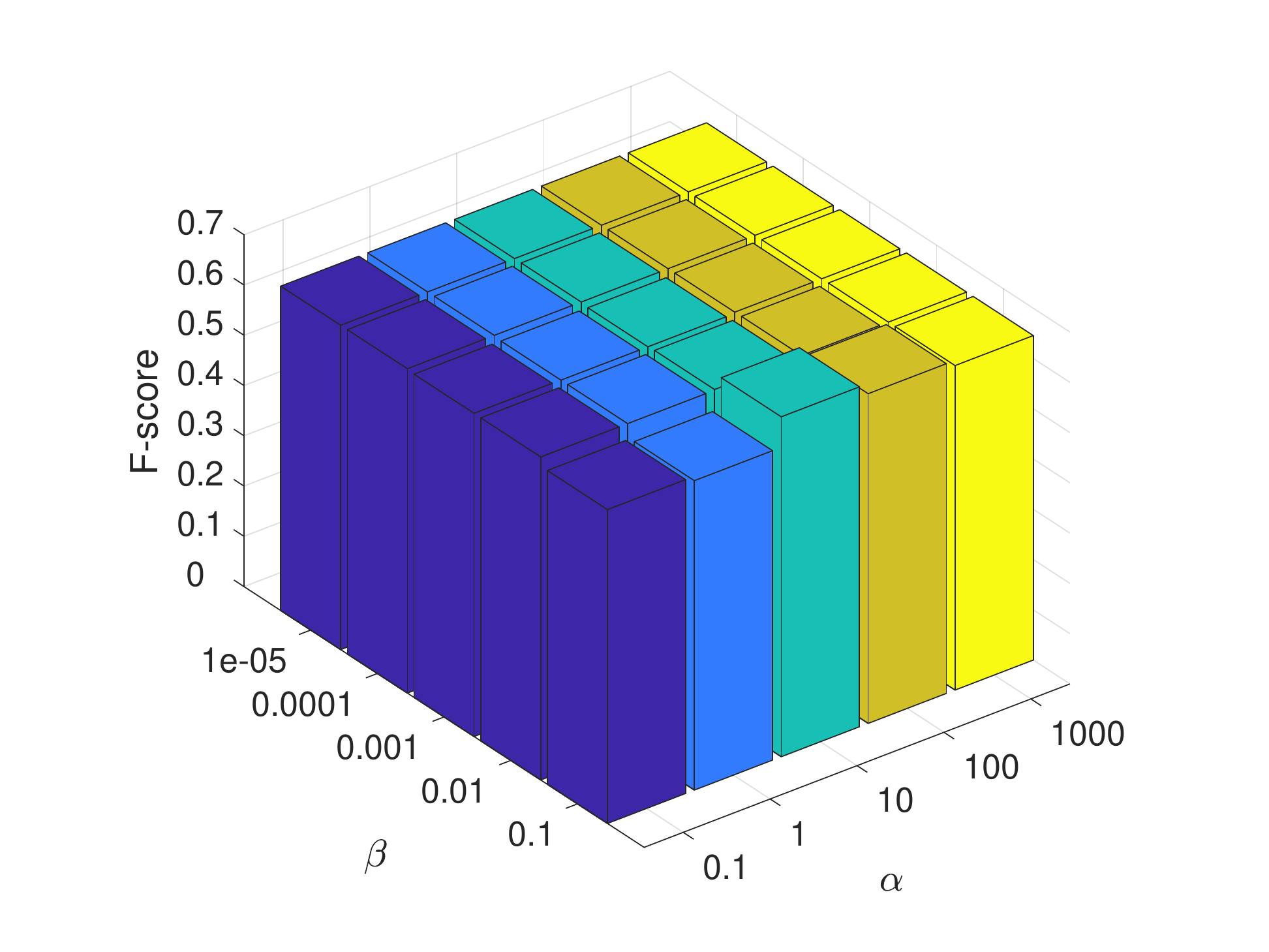}}
\caption{The effect of parameters on the Caltech7 data set.\label{acc}}
\end{figure}

\subsection{Experimental Results}
We repeat each method 10 times and report their mean and standard deviation (std) values. For our proposed method, we only need to implement once since no k-means is involved. The clustering performance on those four data sets is summarized in Tables \ref{bbc}-\ref{handwritten}, respectively. We can observe that our mPAC method achieves the best performance in most cases, which validates the effectiveness of our approach. In general, our method works better than k-means and NMF based techniques. Furthermore, it can be seen that the improvement is remarkable. With respect to graph-based clustering methods, our approach also demonstrates its superiority. In particular, both MVSC and AMGL assume that all graphs produce the same partition, while our method learns one partition for each view and finds the underlying cluster by aligning mechanism.

To visualize the effect of partitions alignment, we implement t-SNE on the clustering results of Handwritten 
Numerals data. As shown in Fig. \ref{partition}, some partitions have a good cluster structure, thus it might be easy to find a good $Y$. On the other hand, although the partition of view 5 is bad, we can still achieve a good solution $Y$. This indicates that our method is reliable to obtain a good clustering since it operates in the partition space. By contrast, previous methods may not consistently provide a good solution. 

\subsection{Sensitivity Analysis}
Taking Caltech7 as an example, we demonstrate the influence of parameters to clustering performance. From Fig.    \ref{acc}, we can observe that our performance is quite stable under a wide range of parameter settings. In particular, it becomes more robust to $\alpha$ and $\beta$ when $\gamma$ increases, which indicates the importance of partition alignment. 
\section{Conclusion}
In this paper, a novel multi-view clustering method is developed. Different from existing approaches, it seeks to integrate multi-view information in partition space. We assume that each partition can be aligned to the consensus clustering through a rotation matrix. Furthermore, graph learning and clustering are performed in a unified framework, so that they can be jointly optimized. The proposed method is validated on four benchmark data sets. 
%
\section*{Acknowledgments}
This paper was in part supported by Grants from the Natural Science Foundation of China (Nos. 61806045 and 61572111), two Fundamental Research Fund for the Central Universities of China (Nos. ZYGX2017KYQD177 and A03017023701012) and a 985 Project of UESTC (No. A1098531023601041) .
\bibliographystyle{named}
\bibliography{ref}

\begin{thebibliography}{}

\bibitem[\protect\citeauthoryear{Cai \bgroup \em et al.\egroup
  }{2013}]{cai2013multi}
Xiao Cai, Feiping Nie, and Heng Huang.
\newblock Multi-view k-means clustering on big data.
\newblock In {\em IJCAI}, pages 2598--2604, 2013.

\bibitem[\protect\citeauthoryear{Chao \bgroup \em et al.\egroup
  }{2017}]{chao2017survey}
Guoqing Chao, Shiliang Sun, and Jinbo Bi.
\newblock A survey on multi-view clustering.
\newblock {\em arXiv preprint arXiv:1712.06246}, 2017.

\bibitem[\protect\citeauthoryear{Chen \bgroup \em et al.\egroup
  }{2013}]{chen2013twkm}
Xiaojun Chen, Xiaofei Xu, Yunming Ye, and Joshua~Zhexue Huang.
\newblock {TW-k-means: Automated Two-level Variable Weighting Clustering
  Algorithm for Multi-view Data}.
\newblock {\em IEEE Transactions on Knowledge and Data Engineering},
  25(4):932--944, 2013.

\bibitem[\protect\citeauthoryear{Elhamifar and
  Vidal}{2013}]{elhamifar2013sparse}
Ehsan Elhamifar and Rene Vidal.
\newblock Sparse subspace clustering: Algorithm, theory, and applications.
\newblock {\em IEEE transactions on pattern analysis and machine intelligence},
  35(11):2765--2781, 2013.

\bibitem[\protect\citeauthoryear{Gao \bgroup \em et al.\egroup
  }{2015}]{gao2015multi}
Hongchang Gao, Feiping Nie, Xuelong Li, and Heng Huang.
\newblock Multi-view subspace clustering.
\newblock In {\em ICCV}, pages 4238--4246, 2015.

\bibitem[\protect\citeauthoryear{Huang \bgroup \em et al.\egroup
  }{2018}]{huang2018self}
Shudong Huang, Zhao Kang, and Zenglin Xu.
\newblock Self-weighted multi-view clustering with soft capped norm.
\newblock {\em Knowledge-Based Systems}, 158:1--8, 2018.

\bibitem[\protect\citeauthoryear{Huang \bgroup \em et al.\egroup
  }{2019}]{huang2019auto}
Shudong Huang, Zhao Kang, Ivor~W Tsang, and Zenglin Xu.
\newblock Auto-weighted multi-view clustering via kernelized graph learning.
\newblock {\em Pattern Recognition}, 88:174--184, 2019.

\bibitem[\protect\citeauthoryear{Jain}{2010}]{jain2010data}
Anil~K Jain.
\newblock Data clustering: 50 years beyond k-means.
\newblock {\em Pattern recognition letters}, 31(8):651--666, 2010.

\bibitem[\protect\citeauthoryear{Kang \bgroup \em et al.\egroup
  }{2017}]{kang2017twin}
Zhao Kang, Chong Peng, and Qiang Cheng.
\newblock Twin learning for similarity and clustering: A unified kernel
  approach.
\newblock In {\em AAAI}, 2017.

\bibitem[\protect\citeauthoryear{Kang \bgroup \em et al.\egroup
  }{2018a}]{kang2018self}
Zhao Kang, Xiao Lu, Jinfeng Yi, and Zenglin Xu.
\newblock Self-weighted multiple kernel learning for graph-based clustering and
  semi-supervised classification.
\newblock In {\em IJCAI}, pages 2312--2318. AAAI Press, 2018.

\bibitem[\protect\citeauthoryear{Kang \bgroup \em et al.\egroup
  }{2018b}]{kang2018unified}
Zhao Kang, Chong Peng, Qiang Cheng, and Zenglin Xu.
\newblock Unified spectral clustering with optimal graph.
\newblock In {\em AAAI}, 2018.

\bibitem[\protect\citeauthoryear{Kang \bgroup \em et al.\egroup
  }{2019a}]{kang2019similarity}
Zhao Kang, Yiwei Lu, Yuanzhang Su, Changsheng Li, and Zenglin Xu.
\newblock Similarity learning via kernel preserving embedding.
\newblock In {\em AAAI}, 2019.

\bibitem[\protect\citeauthoryear{Kang \bgroup \em et al.\egroup
  }{2019b}]{kang2019cyber}
Zhao Kang, Haiqi Pan, Steven~C.H. Hoi, and Zenglin Xu.
\newblock Robust graph learning from noisy data.
\newblock {\em IEEE Transactions on Cybernetics}, pages 1--11, 2019.

\bibitem[\protect\citeauthoryear{Kumar and Daum{\'e}}{2011}]{kumar2011cotrain}
Abhishek Kumar and Hal Daum{\'e}.
\newblock A co-training approach for multi-view spectral clustering.
\newblock In {\em ICML}, pages 393--400, 2011.

\bibitem[\protect\citeauthoryear{Kumar \bgroup \em et al.\egroup
  }{2011}]{kumar2011co}
Abhishek Kumar, Piyush Rai, and Hal Daume.
\newblock Co-regularized multi-view spectral clustering.
\newblock In {\em NIPS}, pages 1413--1421, 2011.

\bibitem[\protect\citeauthoryear{Liu \bgroup \em et al.\egroup
  }{2018}]{liu2018partition}
Hongfu Liu, Zhiqiang Tao, and Yun Fu.
\newblock Partition level constrained clustering.
\newblock {\em IEEE Transactions on Pattern Analysis and Machine Intelligence},
  40(10):2469--2483, Oct 2018.

\bibitem[\protect\citeauthoryear{Ng \bgroup \em et al.\egroup
  }{2002}]{ng2002spectral}
Andrew~Y Ng, Michael~I Jordan, Yair Weiss, et~al.
\newblock On spectral clustering: Analysis and an algorithm.
\newblock {\em NIPS}, 2:849--856, 2002.

\bibitem[\protect\citeauthoryear{Nie \bgroup \em et al.\egroup
  }{2016a}]{nie2016parameter}
Feiping Nie, Jing Li, Xuelong Li, et~al.
\newblock Parameter-free auto-weighted multiple graph learning: A framework for
  multiview clustering and semi-supervised classification.
\newblock In {\em IJCAI}, pages 1881--1887, 2016.

\bibitem[\protect\citeauthoryear{Nie \bgroup \em et al.\egroup
  }{2016b}]{nie2016constrained}
Feiping Nie, Xiaoqian Wang, Michael~I Jordan, and Heng Huang.
\newblock The constrained laplacian rank algorithm for graph-based clustering.
\newblock In {\em AAAI}, 2016.

\bibitem[\protect\citeauthoryear{Nie \bgroup \em et al.\egroup
  }{2018}]{nie2018multiview}
Feiping Nie, Lai Tian, and Xuelong Li.
\newblock Multiview clustering via adaptively weighted procrustes.
\newblock In {\em SIGKDD}, pages 2022--2030. ACM, 2018.

\bibitem[\protect\citeauthoryear{Peng \bgroup \em et al.\egroup
  }{2018}]{peng2018connections}
Xi~Peng, Canyi Lu, Zhang Yi, and Huajin Tang.
\newblock Connections between nuclear-norm and frobenius-norm-based
  representations.
\newblock {\em IEEE transactions on neural networks and learning systems},
  29(1):218--224, 2018.

\bibitem[\protect\citeauthoryear{Sch{\"o}nemann}{1966}]{schonemann1966generalized}
Peter~H Sch{\"o}nemann.
\newblock A generalized solution of the orthogonal procrustes problem.
\newblock {\em Psychometrika}, 31(1):1--10, 1966.

\bibitem[\protect\citeauthoryear{Tzortzis and Likas}{2012}]{tzortzis2012kernel}
Grigorios Tzortzis and Aristidis Likas.
\newblock Kernel-based weighted multi-view clustering.
\newblock In {\em ICDM}, pages 675--684. IEEE, 2012.

\bibitem[\protect\citeauthoryear{Wang \bgroup \em et al.\egroup
  }{2016}]{wang2016iterative}
Yang Wang, Wenjie Zhang, Lin Wu, Xuemin Lin, Meng Fang, and Shirui Pan.
\newblock Iterative views agreement: an iterative low-rank based structured
  optimization method to multi-view spectral clustering.
\newblock In {\em IJCAI}, pages 2153--2159. AAAI Press, 2016.

\bibitem[\protect\citeauthoryear{Wen and Yin}{2013}]{wen2013feasible}
Zaiwen Wen and Wotao Yin.
\newblock A feasible method for optimization with orthogonality constraints.
\newblock {\em Mathematical Programming}, 142(1-2):397--434, 2013.

\bibitem[\protect\citeauthoryear{Yang \bgroup \em et al.\egroup
  }{2018}]{yang2018fast}
Xiaojun Yang, Weizhong Yu, Rong Wang, Guohao Zhang, and Feiping Nie.
\newblock Fast spectral clustering learning with hierarchical bipartite graph
  for large-scale data.
\newblock {\em Pattern Recognition Letters}, 2018.

\bibitem[\protect\citeauthoryear{Zhan \bgroup \em et al.\egroup
  }{2017}]{zhan2017graph}
Kun Zhan, Changqing Zhang, Junpeng Guan, and Junsheng Wang.
\newblock Graph learning for multiview clustering.
\newblock {\em IEEE transactions on cybernetics}, (99):1--9, 2017.

\bibitem[\protect\citeauthoryear{Zhang \bgroup \em et al.\egroup
  }{2017}]{zhang2017latent}
Changqing Zhang, Qinghua Hu, Huazhu Fu, Pengfei Zhu, and Xiaochun Cao.
\newblock Latent multi-view subspace clustering.
\newblock In {\em CVPR}, pages 4279--4287, 2017.

\bibitem[\protect\citeauthoryear{Zong \bgroup \em et al.\egroup
  }{2017}]{zong2017multi}
Linlin Zong, Xianchao Zhang, Long Zhao, Hong Yu, and Qianli Zhao.
\newblock Multi-view clustering via multi-manifold regularized non-negative
  matrix factorization.
\newblock {\em Neural Networks}, 88:74--89, 2017.

\end{thebibliography}
\end{document}